\pdfoutput=1
\documentclass[letterpaper, 10 pt, conference]{ieeeconf}
\IEEEoverridecommandlockouts 
\usepackage{times,multirow}

\usepackage{amsfonts,amssymb,amsmath,booktabs,upgreek}

\usepackage{epsfig,sped,times,wrapfig}
\usepackage{algorithm,url,textcomp,tabularx}
\usepackage{color}
\usepackage{subfigure}
\usepackage{multirow}
\usepackage{soul}
\usepackage[noend]{algorithmic}
\usepackage{verbatim}
\usepackage{amsmath}
\usepackage{tikz}

\usepackage[skip=3pt,size=footnotesize]{caption}
\usepackage{mathtools}
\usepackage{graphicx}
\usepackage{pgfplots} 
\usepackage{diagbox}
\usepackage{rotating}
\usepackage{tabularx}
\usepackage{booktabs}
\usepackage{colortbl}
\graphicspath{{./figures/}}
\setulcolor{green}

\makeatletter
\@addtoreset{algorithm}{section}
\makeatother

\newcolumntype{a}{>{\centering\arraybackslash\columncolor{lightgray!30}}p{0.75cm}}
\newcolumntype{b}{>{\centering\arraybackslash}p{0.75cm}}

\newcommand{\name} {\mbox{RF-Grasp}} 

\newenvironment{Itemize}%
{\begin{itemize}%
\setlength{\itemsep}{0pt}%
\setlength{\topsep}{0pt}%
\setlength{\partopsep}{0pt}%
\setlength{\parskip}{0pt}}%
{\end{itemize}}
{\begin{enumerate}%
\setlength{\itemsep}{0pt}%
\setlength{\topsep}{0pt}%
\setlength{\partopsep}{0pt}%
\setlength{\parskip}{0pt}}%
{\end{enumerate}}
\makeatletter
  \newcommand\figcaption{\def\@captype{figure}\caption}
  \newcommand\tabcaption{\def\@captype{table}\caption}
\makeatother

\newcommand{\textred}[1]{\textcolor{red}{#1}}
\ifx\noeditingmarks\undefined
   \newcommand{\pgwrapper}[2]{\textred{#1: #2}}
\else
   \newcommand{\pgwrapper}[2]{}
\fi

\newcommand{\textblue}[1]{\textcolor{blue}{#1}}
\ifx\noeditingmarks\undefined
   \newcommand{\bpgwrapper}[2]{\textblue{#1: #2}}
\else
   \newcommand{\bpgwrapper}[2]{}
\fi

\begin{document}
\title{Robotic Grasping of Fully-Occluded Objects using RF Perception}



\newcommand{\supsym}[1]{\raisebox{6pt}{{\footnotesize #1}}}

\author{
Tara Boroushaki, Junshan Leng, Ian Clester, Alberto Rodriguez, Fadel Adib \\
Massachusetts Institute of Technology
} 

\maketitle

\begin{sloppypar} 

\begin{abstract}
We present the design, implementation, and evaluation of RF-Grasp, a robotic system that can grasp fully-occluded objects in unknown and unstructured environments. 
Unlike prior systems that are constrained by the line-of-sight perception of vision and infrared sensors, RF-Grasp employs RF (Radio Frequency) perception to identify and locate target objects \textit{through} occlusions, and perform efficient exploration and complex manipulation tasks in non-line-of-sight settings.

RF-Grasp relies on an eye-in-hand camera and batteryless RFID tags attached to objects of interest. It introduces two main innovations: (1) an RF-visual servoing controller that uses the RFID’s location to selectively explore the environment and plan an efficient trajectory toward an occluded target, and (2) an RF-visual deep reinforcement learning network that can learn and execute efficient, complex policies for decluttering and grasping.

We implemented and evaluated an end-to-end physical prototype of RF-Grasp. We demonstrate it improves success rate and efficiency by up to 40-50\% over a state-of-the-art baseline. We also demonstrate \name\ in novel tasks such mechanical search of fully-occluded objects behind obstacles, opening up new possibilities for robotic manipulation. Qualitative results (videos) available at \textcolor{blue}{\footnotesize{\url{rfgrasp.media.mit.edu}}}

\end{abstract}



\vspace{-0.1in}
\section{Introduction}
\vspace{-0.05in}

Mechanical search is a fundamental problem in robotics~\cite{IROS2019Xray,ICRA2019search,objectfinding,graspinvisible}. It refers to the task of searching for and retrieving a partially or fully-occluded target object. This problem arises frequently in unstructured environments such as warehouses, hospitals, agile manufacturing plants, and homes. For example, a warehouse robot may need to retrieve an e-commerce customer’s desired item from under a pile. Similarly, a robot may need to retrieve a desired tool (e.g., screwdriver) from behind an obstacle to perform a complex task such as furniture assembly~\cite{ikeabot}.
%
%

To address this problem, the past few years have seen significant advances in learning models that can either recognize target objects through partial occlusions or actively explore the environment, searching for the object of interest. Recent proposals have also considered the geometry of obstacles or pile~\cite{IROS2019Xray,ICRA2019search,BerkeleyGraspOcclusion,afforfances1,planningclutter}, demonstrating remarkable results in efficiently exploring and decluttering the environment. 

However, existing mechanical search systems are inherently constrained by their vision sensors, which can only perceive objects in their direct line-of-sight. If the object of interest is behind an obstacle, they need to actively explore the environment searching for it, a process that can be very expensive and often fails~\cite{BerkeleyGraspOcclusion}. Moreover, these systems are typically limited to a single pile or pick-up bin~\cite{IROS2019Xray,objectfinding}, and cannot generalize to  mechanical search problems with multiple piles or multiple obstacles. They also cannot perform tasks like prioritized sorting, where a robot needs to retrieve \textit{all} objects belonging to a \textit{specific} class (e.g., all plastic bottles from a box) and then declare task completion.

\begin{figure}[t]
	\centering
	\setlength{\tabcolsep}{0pt}
	\begin{tabular}{cc}
	\multicolumn{2}{c}{
		\epsfig{file=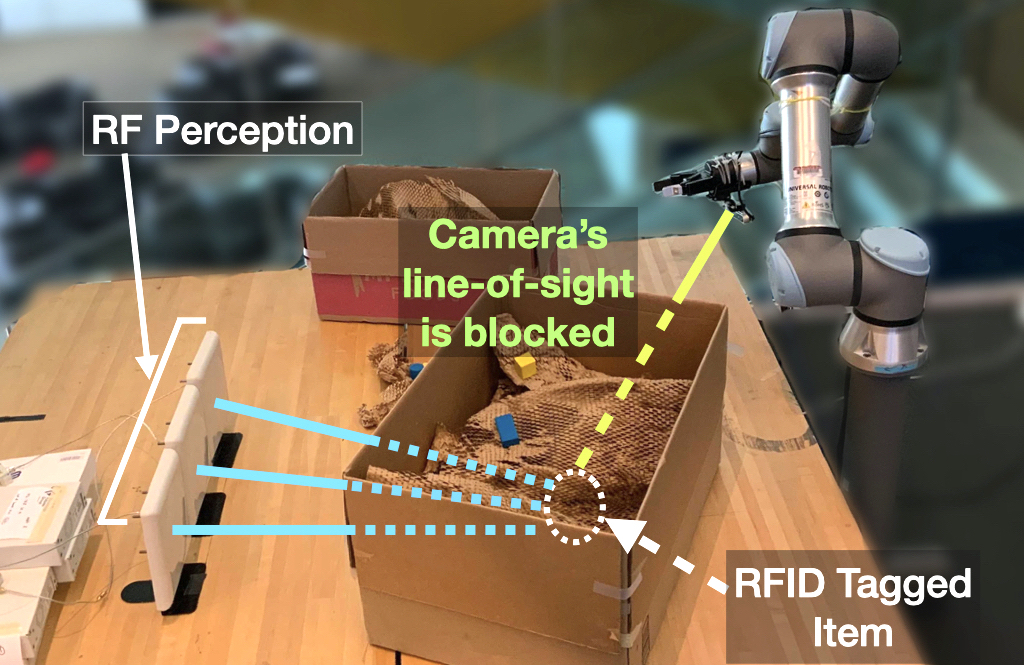,width=0.48\textwidth}}
		\vspace{-0.05in}
		\\
	\multicolumn{2}{c}{
		\footnotesize{(a) Target object is fully-occluded in an unstructured environment.}}
		\\
		\epsfig{file=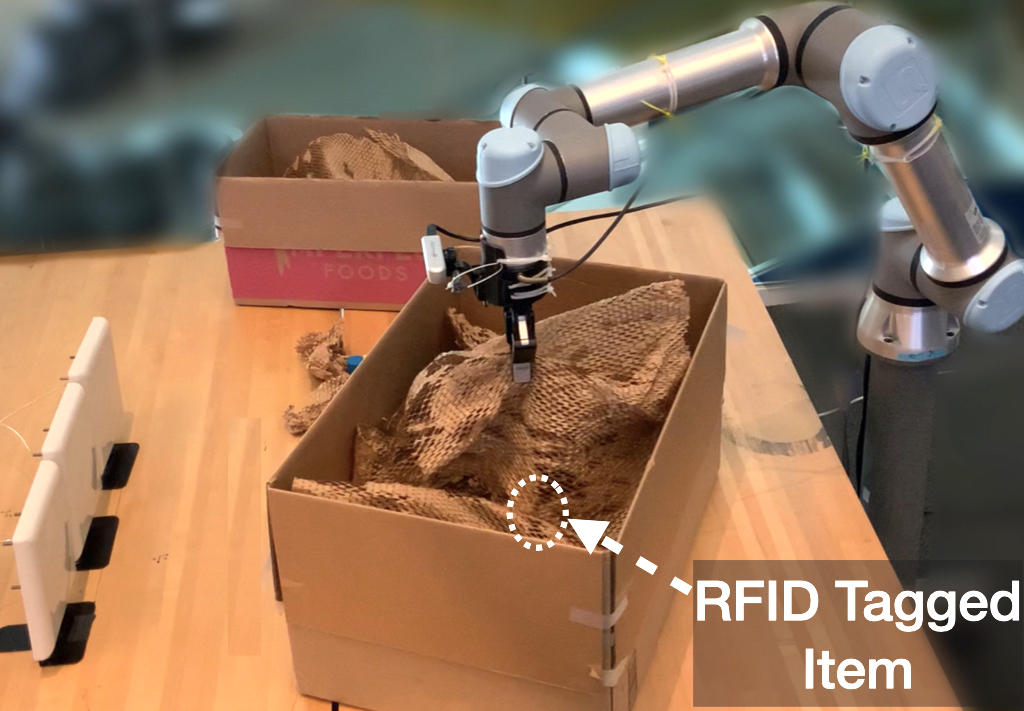,width=0.235\textwidth}
		\vspace{-0.02in}
		&
		\epsfig{file=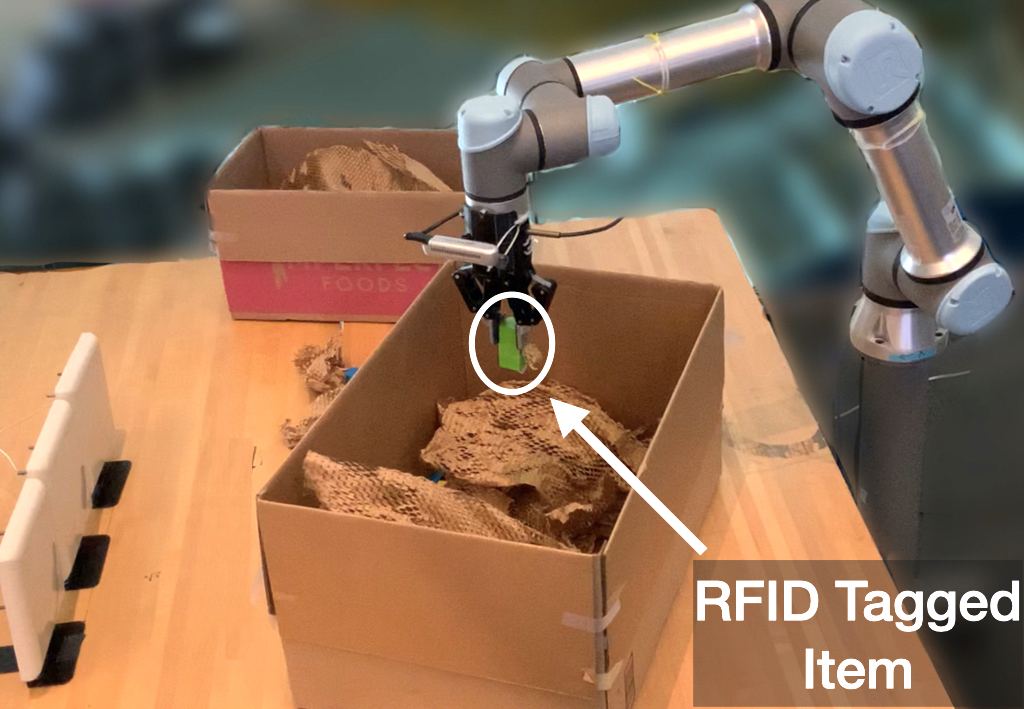,width=0.235\textwidth}
		\vspace{-0.02in}
		\\
		\footnotesize{(b) Decluttering}
		& 
		\footnotesize{(c) Successful Grasping}
\end{tabular}
\vspace{-0.05in}
\caption {\textbf{RF-Grasp grasping fully-occluded objects.} (a) Even though the camera and RF sensor have no line-of-sight to the object, \name\ uses RF perception to identify and locate the target object through occlusions. (b) It employs \textit{RF-visual servoing} and \textit{RF-visual manipulation} to efficiently maneuver toward the object and declutter its  vicinity. (c) Successful pick-up.}
\label{fig:intro-fig}
\vspace{-0.15in}
\end{figure}

In this paper, we draw on recent advances in RF (Radio Frequency) perception~\cite{TurboTrack,RFCapture} to enable novel and highly-efficient mechanical search tasks in unstructured environments. Unlike visible light and infrared, RF signals can traverse everyday occlusions like cardboard boxes, wooden dividers (and walls), opaque plastic covers, and colored glass. This “see through” capability enables a robot to perceive objects tagged with passive 3-cent RF stickers (called RFIDs), 
even when they are fully occluded from its vision sensor. RFID systems can read and uniquely identify
hundreds of tags per second from up to 9~m and through occlusions~\cite{Imping}. 

The key challenge with RF perception is that unlike vision, it cannot produce high-resolution images with pixel-wise precision. Rather, it only obtains the 3D tag location with centimeter-scale precision~\cite{TurboTrack}. Moreover, because standard (visual) occlusions are transparent to RF signals, a robot can neither perceive them using RF sensing nor reason about the visual exploration, collision avoidance, and decluttering steps that may be necessary prior to grasping the target object.


We present RF-Grasp (shown in Fig.~\ref{fig:intro-fig}), the first robotic system that fuses RF and vision information (RF+RGB-D) to enable efficient and novel mechanical search tasks across line-of-sight, non-line-of-sight, and highly cluttered environments. This paper provides three main contributions:
 \begin{Itemize}
    \item It presents the first system that bridges RF and vision perception \textit{(RF+RGB-D)} to enable mechanical search and extended mechanical search.
    \item It introduces two novel primitives: (1) \textit{RF-visual servoing}, a novel controller that performs RF-guided active exploration and navigation to avoid obstacles and maneuver toward the target object, and (2) \textit{RF-visual grasping}, a model-free deep reinforcement learning network that employs RF-based attention to learn and execute efficient and complex policies for decluttering and grasping.
    \item It presents an end-to-end prototype implementation and evaluation of \name. The implementation is built using a UR5e robot, Intel RealSense D415 depth camera, and a customized RF localization system on software radios. The system is evaluated in over 100 physical experiments and compared to a baseline that combines two prior systems~\cite{BerkeleyGraspOcclusion,graspinvisible}. The evaluation demonstrates that \name\ improves success rate by up to 40\% and efficiency by up to 50\% in challenging environments. Moreover, it demonstrates that \name\ can perform novel mechanical search tasks of fully-occluded objects across settings with obstacles, multiple piles, and multiple target objects.
\end{Itemize}

In comparison to vision-only systems, \name's design requires target object(s) tagged with RFIDs. However, we believe that given the widespread adoption of RFIDs by many industries (with tens of billions deployed annually~\cite{RFIDmarket}), the system can already have significant practical impact. 
We also hope this work motivates further research bridging RF and vision for novel (and more efficient) robotic tasks.

\vspace{-0.05in}
\section{Background \& Related Work}
\vspace{-0.05in}

\begin{figure}[t]
	\begin{center}
		\includegraphics[width =0.42\textwidth]{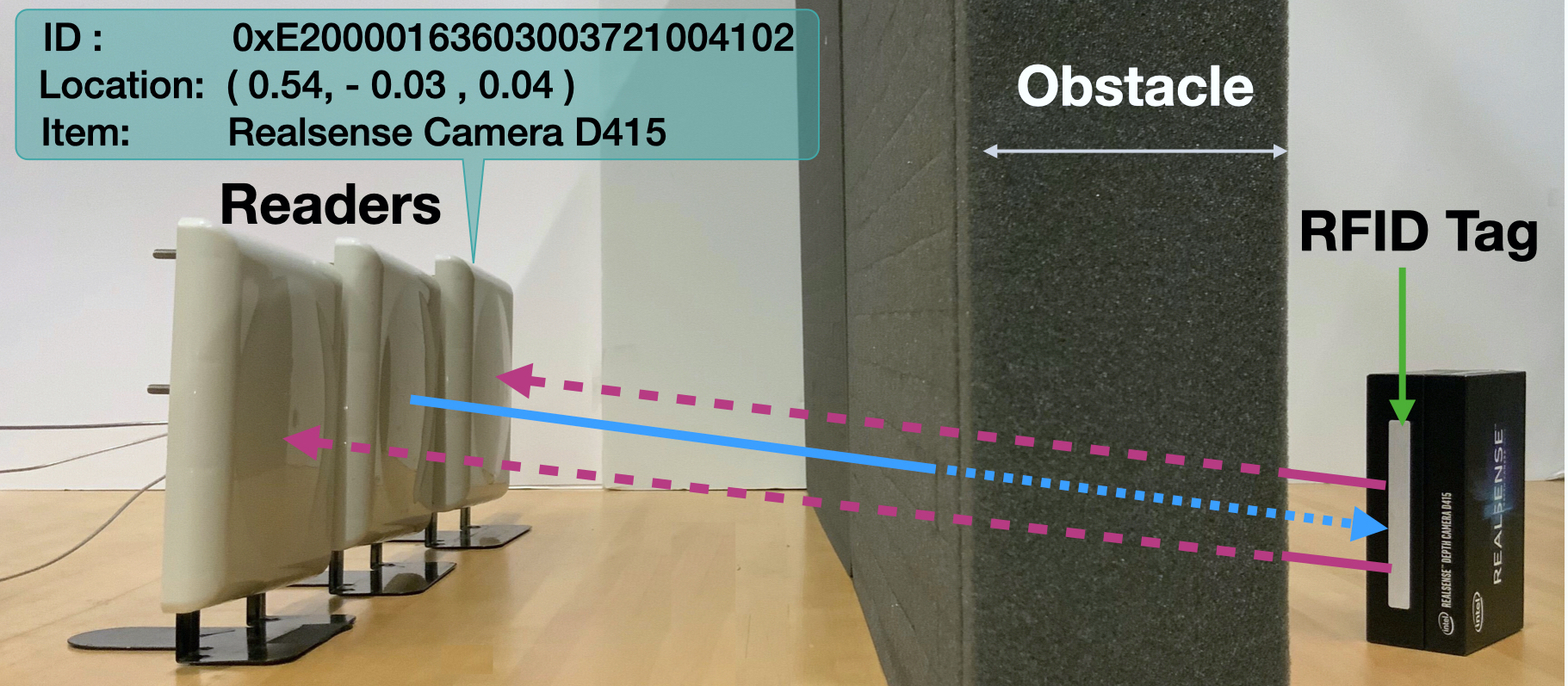}
		\caption{\textbf{RFID-based Perception}. \name\ uses a customized reader that can identify and accurately localize RFID-tagged objects through occlusions.}
		\label{fig:rfid}
	\end{center}
	\vspace{-0.17in}
\end{figure}

\noindent
\textbf{RFIDs and their Applications in Robotics.}
Radio Frequency IDentification (RFID) is a mature technology, widely adopted by many industries as barcode replacement 
in retail, manufacturing, and warehousing~\cite{RFIDmarket}. 
%
Recent years have seen significant advances in RFID localization technologies~\cite{PinIt,RFCompass,MobiTag}, which not only use the tags for identification, but also locate them with centimeter-scale precision even in cluttered and occluded settings~\cite{RFind,TurboTrack}, as in Fig.~\ref{fig:rfid}.

\begin{figure*}[thp]
	\begin{center}
		\includegraphics[width =\textwidth]{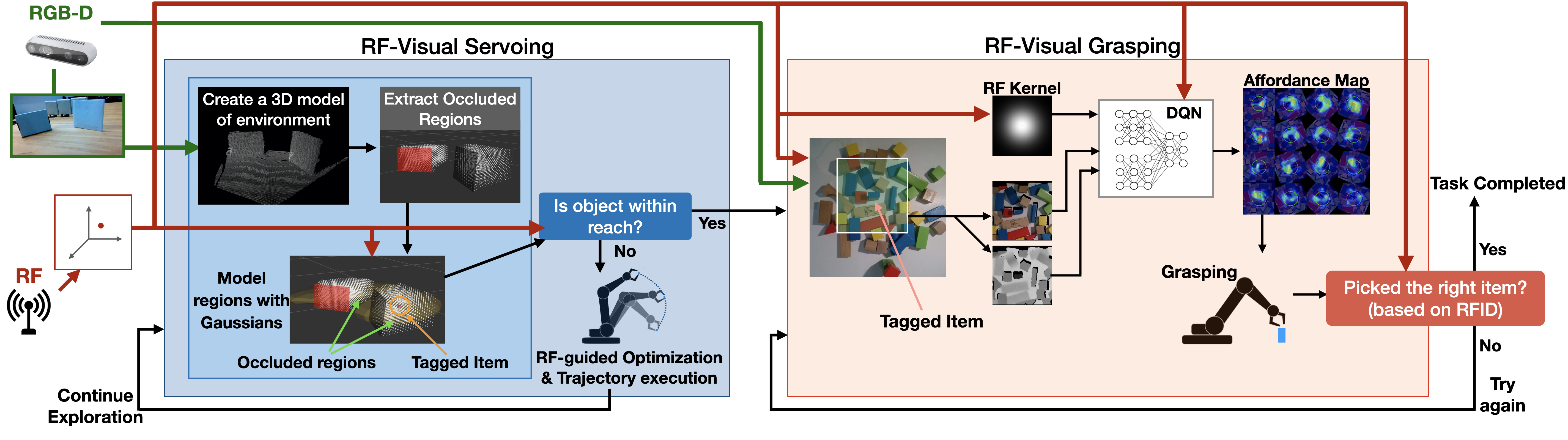}
		\vspace{-0.22in}
		\caption{\textbf{System Overview.} \name\ uses RGB-D (green arrows) to create a 3D model of the environment and extract occluded regions, and it plugs the location of the RFID-tagged object (red arrows) into it. If the object is behind obstacles, it performs RF-guided active exploration and trajectory optimization to obtain a better view of the object by maneuvering around obstacles. Once a stopping criterion is met, it proceeds to grasping. Here, it applies RF-based attention to the RGB-D information and relies on a model-free deep-reinforcement network to discover optimal grasping affordances. \name\ uses the RFID's locationto close the loop on the grasping task and determine whether it has been successful or whether it needs to make another grasping attempt.}
		\label{fig:overview}
	\end{center}
	\vspace{-0.17in}
\end{figure*}

Prior work leveraged RFIDs as location markers for robotic navigation~\cite{navigation1,navigation2,navigation3} and to guide mobile robots toward grasping~\cite{RFCompass,T25,T28,T31,MobiTag,RFIDfusion}. But because occlusions are transparent to RF, these systems could not perceive them to declutter or maneuver around them. 
In contrast, \name\ demonstrates, for the first time, how RF-visual fusion enables complex tasks like mechanical search.



\vspace{0.03in}
\noindent
\textbf{Mechanical Search and Grasping in Clutter.}
%
Object manipulation in cluttered environments has received significant attention and achieved remarkable success via supervised and unsupervised methods~\cite{PushGrasp,MVPclutter,rpp,graspocclusion}.
Our work is motivated by a similar desire to grasp in clutter and builds on this line of work but focuses on the problem of grasping a specific target object rather than \textit{any} object or \textit{all} objects.

Recognizing and locating target objects in occlusions has also received much attention~\cite{semantic1}. Various techniques have been proposed including perceptual completion~\cite{occlusionaware,price2019inferring} and active/interactive perception where a camera is moved to more desirable vantage points for recognizing objects~\cite{activeperception1,activeperception2,interactiveperception}. In contrast to this past work, which requires a (partial) line-of-sight to an object to recognize it, our work uses RF perception to directly identify and locate objects through occlusions, and without requiring any prior training.

\name\ is most related to recent prior work on mechanical search of partially or fully occluded target objects~\cite{BerkeleyGraspOcclusion,IROS2019Xray,ICRA2019search,T36}; this includes both one-shot and multi-step procedures for search and retrieval~\cite{graspinvisible,objectfinding,T37,T38,T39}. Unlike \name, this prior work first needs to search for the object to identify its location, which does not scale well with the number and size of piles, or the number of target objects; in contrast, by exploiting RF perception, \name\ can recognize and locate RFID-tagged objects to perform highly efficient active exploration and manipulation.


\vspace{-0.02in}
\section{System Overview}
\vspace{-0.07in}

We  consider  a  generalized  mechanical  search  problem where  a  robot  needs  to  extract a target object in an unstructured environment. The object may be in line-of-sight or non-line-of-sight; it may be behind occlusions and/or under a pile, and the environment may have additional occlusions, piles, and clutter, similar to Fig.~\ref{fig:intro-fig}. Moreover, the robot may need to extract all target objects from a semantic class. 

We assume that each target object (but not necessarily other objects) is tagged with a UHF RFID and kinematically reachable from a robotic arm on a fixed base. We also assume the environment is static. 
The robot is equipped with an eye-in-hand camera, mounted on a 6-DOF manipulator, which starts from a random initial location and orientation. The robot is aided by a fixed-mount RF perception module in the form of an RFID micro-location sensor with multiple antennas. The robot knows the target object(s) RFID number but no additional information about its geometry or location.

\name’s objective is to extract the target(s) from the environment using the shortest travel distance and the minimum number of grasp attempts. It starts by querying the RFIDs in the environment and using its RF perception module to identify them and compute their 3D locations, even if they are behind occlusions~\cite{TurboTrack}.
It divides the mechanical search problem into two sub-problems as shown in Fig.~\ref{fig:overview} and addresses each of them using a separate subcomponent:
\vspace{-0.04in}
\begin{itemize}
\item \textbf{RF-Visual Servoing:} The first aims to maneuver the robotic arm toward the target object. It uses RGB-D images to create a 3D model of the environment
and fuses the RFID’s location into it. It then performs RF-guided active exploration and trajectory optimization to efficiently maneuver around obstacles toward the object. Exploration stops when it can grasp the object or declutter its vicinity.

\item \textbf{RF-Visual Grasping:} \name’s second sub-component is a model free deep-reinforcement learning network that aims to identify optimal grasping affordances from RGB-D, using the RFID’s location as an attention mechanism. The robot attempts to grasp and pick up the target object, and stops once the RF perception module determines that the RFID’s location has moved with the end-effector. 
\end{itemize}

\vspace{-0.05in}
\vspace{-0.02in}
\section{RF-Visual Servoing}\label{sec:servoing}
\vspace{-0.05in}

Given the RFID’s location and an RGB-D image, \name\ needs to maneuver the robotic arm toward the target object. A key difficulty in this process is that the environment is not known a priori and the direct path may be occluded by obstacles. Below, we describe how \name\ actively explores the environment as it tries to determine an optimal path around obstacles toward the target object. 


\vspace{-0.05in}
\subsection{Problem Definition}
\vspace{-0.05in}
We frame the servoing problem as a Partially Observable Markov Decision Process (POMDP) where the robot needs to efficiently explore the environment while minimizing the trajectory toward the object of interest. The state of the environment $\mathcal{(S)}$ consists of the robot joint state ($\mathbf{x}_t^R\in\mathbb{R}^6$), RFID location ($\mathbf{p}=(x_p,y_p,z_p)$), the obstacles, occlusions, and other objects. The  control  signal, $\mathbf{u}_t\in\mathbb{R}^6$, is applied to  the  joint  state, and changes the robot pose. The observations $(\Omega)$ consist of the joint state, the RFID location, and RGB-D data from the wrist-mounted camera. The problem is partially observable because the robot has no prior knowledge of (nor observes)  the entire 3D workspace.



\begin{figure}[t] 
\centering
\epsfig{file=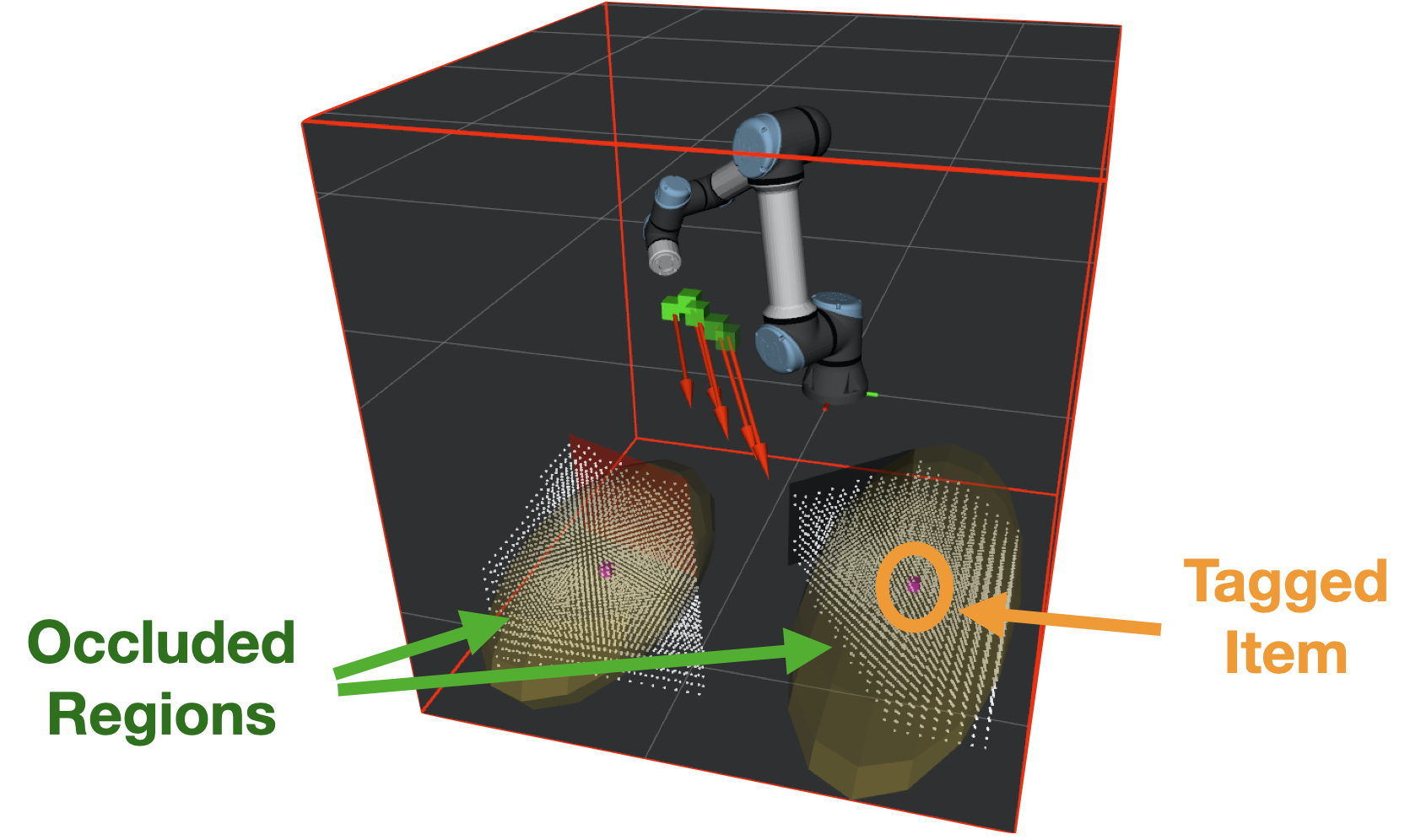,width=2.45in}
\vspace{-0.06in}
\captionof{figure}{\footnotesize \textbf{RF-Visual Servoing.} \name\ encodes occluded regions as Gaussians and biases its active exploration toward the RFID-tagged target item.}
\label{fig:MoG}
\vspace{-0.125in}
\end{figure}

\noindent
\textbf{Modeling Environmental Uncertainties.}
Similar to past work~\cite{BerkeleyGraspOcclusion}, \name\ encodes uncertainty using a mixture of Gaussians. Each occluded region is modeled as a 3D Gaussian as shown in Fig.~\ref{fig:MoG}. The mean and covariance of the m-th Gaussian are denoted $(\mathbf{x}^{m},\mathbf{\Sigma}_0^{m})$ at $t=0$. The environment is assumed to be static; hence, the means remain the same over the planning horizon, but the covariances $\mathbf{\Sigma}_t^{m}$ get updated as the system explores the environment.  

\noindent
\textbf{RF-Biased Objective Function.}
To efficiently maneuver toward the object, \name\ aims to minimize its trajectory (control effort) while minimizing its uncertainty of the surrounding environment. Mathematically, the cost at time $t$ is:
\vspace{-0.15in}
{
\setlength{\abovedisplayskip}{0pt}
\setlength{\belowdisplayskip}{0pt}
\setlength{\abovedisplayshortskip}{0pt}
\setlength{\belowdisplayshortskip}{0pt}
\begin{align}
C_{t=0:T} (\mathbf{x}_t^{R}, \Sigma_t^{1:M},\mathbf{u}_t) &= \alpha||\mathbf{u}_t||_2^2+\sum\limits_{m=1}^M\beta_t^m\text{tr}(\Sigma_t^m)
\end{align}
}

\vspace{-0.07in}
\noindent
where 
$M$ is the total number of occluded regions, tr is trace, $\alpha$ and $\beta_t^m$ are scalar weighting parameters. 

To bias the controller to explore the occluded region surrounding the RFID, we set that region's corresponding weight, $\beta_t^1$, to be significantly larger than others. Moreover, to give the robot more flexibility to explore in the beginning, we start with a lower $\beta_t^1$ and increase it over time. 

Given the above cost function, we can now formulate the trajectory optimization problem as a minimization function over the planning horizon $T$ as follows:
\vspace{-0.1in}
\begin{equation*}\label{eq:obj1}
\begin{array}{rllr}
\displaystyle \min_{\mathbf{x}_{0:T}^{R},\mathbf{u}_{0:T}} & \multicolumn{3}{l}{\mathbb{E}[ \sum\limits_{t=0}^{T} C_t (\mathbf{x}_t^{R}, \Sigma_t^{1:M},\mathbf{u}_t)]} \\
\textrm{s.t.} & \mathbf{x}_{t+1}^{R} =  \mathbf{f}(\mathbf{x}_{t}^{R},\mathbf{u_t},0),\  \mathbf{x}_{t}^{R} \in \mathbb{X}_{\text{feasible}},\  \mathbf{x}_{0}^{R} =\mathbf{x}_{\text{init}}^{R} & & \\
 &  \mathbf{u}_{t}^{R} \in \mathbb{U}_{\text{feasible}}, \ \mathbf{u}_T=0 
\end{array}
\end{equation*}

\vspace{-0.07in}
\noindent
where  $\mathbb{X}_{\text{feasible}}$ and $\mathbb{U}_{\text{feasible}}$ represent the set of feasible joint states and control signals of the robot arm, $\mathbf{x}_{\text{init}}^{R}$ is the initial joint state of the robot. The dynamics model for the robot is given by differentiable and stochastic function  $\mathbf{x}_{t+1}^{R} = f(\mathbf{x}_{t}^R,\mathbf{u}_t,\mathbf{q}_t), \mathbf{q}_t \sim N(0,I)$ where $\mathbf{q}_t$ is the dynamics noise.

\vspace{-0.03in}
\subsection{RF-Guided Trajectory Optimization}
\vspace{-0.03in}
\name's approach for solving the above problem follows prior work in Gaussian Belief Space Planning (GBSP), including modeling the environment using a 3D voxel grid map,\footnote{It is stored in a TSDF (Truncated Signed Distance Function) volume.} extracting frontiers and occluded regions, dealing with discontinuities in the RGB-D observation model, modeling the observation and dynamics using Gaussians, and propagating beliefs using Extended Kalman Filter (EKF). We refer the interested reader to prior work for details~\cite{BerkeleyGraspOcclusion}, and focus below on two unique features of our solution, aside from the RF-biased objective function described above:

\vspace{0.03in}
\noindent
\textbf{(a) RF-based Initial Trajectory.} To aid the optimization solver and enable faster convergence, we seed the optimization function with a straight-line initial trajectory in Cartesian space from the end-effector to the RFID location.

\vspace{0.03in}
\noindent
\textbf{(b) Exploration Termination Criteria.} In principle, \name\ should stop exploring when it determines that no major obstacles remain, and it can proceed to grasping. But, such reasoning is challenging because the target may still be occluded by distractor objects (e.g., under a pile), and \name\ needs to declutter before grasping.

We formulate the exploration termination criteria as a function of the uncertainty region around the target object. Such uncertainty is encoded both in the covariance of the 3D Gaussian around the object $\Sigma^1$ and in visibility of the voxels $\upsilon$ in the vicinity of $\nu$ the target object's location $\mathbf{p}$. Mathematically, the termination criteria can be expressed as:
\vspace{-0.175in}
\begin{equation}\label{eq:criterion}
\text{trace}(\Sigma^1) < \rho_{\Sigma} \quad \text{or} \quad \sum_{\footnotesize{\upsilon \in \nu(\mathbf{p})}} \digamma(\upsilon)/|\nu(\mathbf{p})| > \rho_{\upsilon}
\end{equation}

\vspace{-0.125in}
\noindent 
where $\digamma(\upsilon)=1$ if voxel $\upsilon$ has been seen by camera and $0$ otherwise. The criteria imply the uncertainty around the object is smaller than threshold $\rho_{\sigma}$, or the visible fraction of the region around the RFID is larger than threshold $\rho_{\upsilon}$.\footnote{In our implementation, $\rho_{\Sigma} = 0.005$ and $\rho_{\upsilon}=0.1$, and the vicinity, $\nu(\mathbf{p})$, is set to a $5\text{cm}\times5\text{cm}\times10\text{cm}$ cube centered at RFID location. } 

The combination of the above criteria is necessary to deal with the diversity of clutter scenarios. For example, when the target item is under a cover that occludes a large region, the trace of covariance won't be below $\rho_\Sigma$. However, enough voxels in the vicinity of the item will be visible to meet the second criterion and \name\ will proceed to grasping.

Finally, it is worth noting that \name's active exploration formulation's objective function pushes the robot end-effector away from collisions. This is because if the camera is too close to a large occlusion, the covariances $\Sigma^m_t$ become larger, thus penalizing the expected cost and biasing the optimal trajectory away from the large obstruction.
\vspace{-0.02in}
\section{Radio-Visual Learning of Grasping
Policies}\label{sec:grasping}
\vspace{-0.05in}


The above primitive enables \name\ to intelligently explore the environment and servo the robotic arm around obstacles, closer to the object of interest, but the robot still needs to grasp it. 
Below, we describe how \name\ exploits RF-based attention to learn efficient grasping policies. 

\vspace{-0.1in}
\subsection{The Grasping Sub-problem}
\vspace{-0.05in}
We formulate the grasping problem as a Markov Decision Process. Here, the action $a_t$ is grasping with a parallel jaw gripper at position $\mathbf{g_t}= (x_{a_t},y_{a_t},z_{a_t})$ with gripper rotation of $\theta_{a_t}$. The goal is to learn the optimal policy $\pi^*$ to grasp the target item (directly or by manipulating the environment).

This can be cast as a deep reinforcement learning problem where the robot aims to maximize the future reward (by grasping the target object). Similar to prior work on unsupervised grasping~\cite{PushGrasp}, we use a Deep Q-Network (DQN). 

\vspace{-0.1in}
\subsection{RF-based Attention and Rewards}
\vspace{-0.05in}

\begin{figure}
	\begin{center}
		\includegraphics[width =0.48\textwidth]{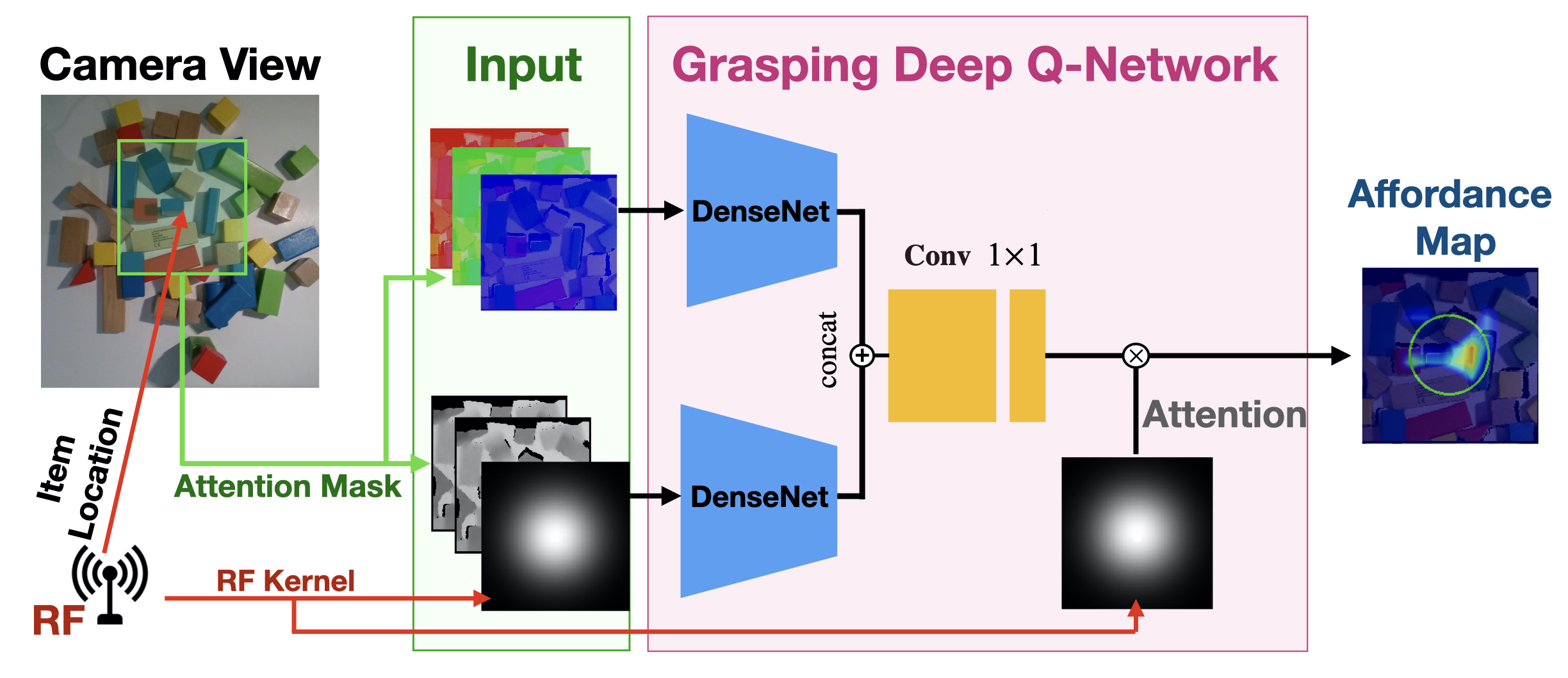}
		\vspace{-0.2in}
		\caption{\textbf{RF-Visual Grasping.} Using the RFID's location, \name\ crops out the RGB-D data near the tagged object location, and feeds the cropped RGB-D data and RFID location probability as input to the Deep-NN that outputs the Q-value estimations. The rotated inputs in 16 different directions are separately fed to DNN to estimate Q-values of different grasping rotations.}
		\label{fig:net}
	\end{center}
	\vspace{-0.225in}
\end{figure}

\name\ trains a deep reinforcement learning network in simulation, using RF information in the reward and attention mechanisms. Fig.~\ref{fig:net} shows the overall architecture, consisting of feed-forward fully convolutional networks. The networks takes RGB-D and RFID position as input, and output pixel-wise map of Q values. The optimal policy is selected as the one with the highest Q across the output affordance map. 

\noindent
\textbf{Spatio-Temporal Reward Function.}
We construct a spatio-temporal reward function to encourage the robot to grasp the target item or those in its near vicinity. The reward function $r(s_t, s_{t+1}, a_t)$ is 1 if the robot grasps the target object;  $\max{(\frac{\varrho}{||\mathbf{p}_t - \mathbf{g}_t||},1)}$ if it grasps another item; and 0 otherwise. $\varrho$ is chosen such that the maximum reward is given to any grasp point within the resolution of RFID positioning.\footnote{In our implementation, $\varrho$ is set to 0.007.} Since RF perception tracks the RFID's location, it can determine whenever the grasp is successful (it is picked up).

\noindent
\textbf{RF-based Attention.} 
\name\ incorporates RF-based attention through two main strategies (and at three layers):
\begin{itemize}
    \item \textit{RF-based Binary Mask.} The RGB and depth heightmaps\footnote{Heightmaps are computed by extracting 3D point cloud from RGB-D images and projecting the images orthographically, parallel to the table top.
    } are cropped to a square around the RFID's location.\footnote{The square is 11cm$\times$11cm in our implementation.} This pre-processing attention mechanism allows the network to focus on the vicinity of the target object and compute the affordance map with higher resolution and/or less computational complexity. 
    \item \textit{RF Kernel.} The RFID location is also used to construct an RF kernel, a 2D Gaussian centered around $\mathbf{p}$, whose standard deviation accounts for RF localization errors. The kernel is fed to the network, and is multiplied by DQN's last layer to compute the final affordance map. This increases the probability of grasping the target item. 
\end{itemize}

\name\ uses the above RF attention mechanisms to extend a state-of-the-art deep-reinforcement learning grasping network~\cite{PushGrasp}, as shown in Fig.~\ref{fig:net}. It consists of two 121-layer DenseNets: the first takes as input three channels of cropped RGB heightmaps; the second's input is the RF kernel plus two copies of cropped depth heightmap. To discover optimal grasping affordances, the inputs are rotated in 16 directions and fed separately to the network. Then, the outputs of both streams are concatenated along channels. Two convolution layers with kernel size of $1\times1$ come after, producing the 2D map of the Q function estimation. The output contains 16 maps for grasp, from which \name\ chooses the position and rotation with highest probability of success.

\vspace{0.05in}
\noindent
\textbf{Training Details.}
The DenseNet initial weights were taken from the pre-trained model in \cite{PushGrasp}, and fine-tuned by training for 500 iterations in simulation. The gradient is only backpropagated through the pixel of affordance map that we executed grasping according to its predicted value. We use Huber loss function and stochastic gradient descent with learning rates  $10^{-4}$, momentum  $0.9$, and weight decay $2^{-5}$ for training. The reward discount factor is 0.2.

In training, we used prioritized experience replay~\cite{prioritized} to improve sample efficiency and stability. 
We define threshold $\rho_i=0.05 + \min(0.05\times\text{\#iteration},4)$, and performed stochastic rank-based prioritization among experiences with rewards $\ge\rho_i$. Prioritization is estimated by a power-law distribution.

\vspace{-0.07in}
\section{Implementation \& Evaluation}
\vspace{-0.08in}
\noindent
\textbf{Physical Prototype.}
Our setup consists of a UR5e robot arm, 2F-85 Robotiq gripper, and an Intel Real-Sense D415 depth camera mounted on the gripper. The RF perception module is implemented as an RFID localization system on USRP software radios (we refer the reader to~\cite{TurboTrack} for implementation details). The RFID localization system is set up on a table in front of the robot arm 16cm below the robot base level. The robot is connected through Ethernet to a PC that runs Ubuntu 16.04 and has an Intel Core i9-9900K processor; RTX 2080 Ti, 11 GB graphic card; and 32 GB RAM. We also used a USB to RS485 interface to control the gripper from the PC.

The robot is controlled using Universal Robots ROS Driver on ROS kinetic. We used Moveit!~\cite{moveit} and OMPL~\cite{OMPL} for planning scene and inverse kinematic solver. TSDF volume is created using Yak~\cite{yak}. We used PCL~\cite{PCL} for extracting clusters and occlusion from TSDF volume. To solve the SQP for trajectory optimization, we used FORCES~\cite{forces}. The code is implemented in both C++ and Python.
Objects of interest are tagged with off-the-shelf UHF RFIDs (e.g., Alien tag~\cite{Alien}), whose dimensions typically range from 1-12~cm.

\noindent
\textbf{Simulation.}
We built a full simulation for \name\ by combining three environments: 1) VREP~\cite{vrep} simulates the depth camera visual data, and robot and gripper's physical interactions with objects in the environment. We used Bullet Physics 2.83 as the physics engine. 2) Gazebo~\cite{gazebo} predicts the robot state. This module takes into account gravity, inertia, and crashing into the ground and other obstacles. 3) Rviz~\cite{rviz} visualizes the robot trajectory, obstacles in the environment, occluded areas, and their mean and co-variances. In the simulated system, we used the VREP remote API to get each object's location to simulate the RFID localization system. Note that only VREP was used to train the DQN, while all three 
were used for RF-Visual servoing.

\noindent
\textbf{Baseline.}
Since prior work does not deal with the generalized mechanical search problem where the target objects are both behind an occlusion \textit{and} dense clutter, we built a baseline by combining two state-of-the-art past systems. The first performs active exploration and trajectory optimization using GBSP~\cite{BerkeleyGraspOcclusion}, but without RF-biasing and guidance. 
The second is a DQN with color-based attention~\cite{graspinvisible} which estimates the location of the item using a unique color. Without loss of generality, the baseline aims to grasp a green item in a workspace with non-green objects, and it switches from exploration to grasping when it detects more than 100 green pixels in its field of view. For fairness to the baseline, we only compare it to our system in scenarios where the object is only partially occluded, but not fully occluded. 

\noindent
\textbf{Evaluation Metrics.}\label{sec:eval}
The robot workspace is atop a table with dimensions of 0.8m$\times$1.2m. 
We consider three metrics: \textit{1) Average Traveled Distance:} the distance that the robot's end-effector moves until grasping the target item. \textit{2) Task Completion Rate:} the percentage of trials that successfully grasped the tagged item (or green item for baseline) before 10 grasping attempts and 5 meter of traveled distance in the exploration phase. 
\textit{3) Grasping Efficiency:} defined as $\frac{\text{\# successful grasps}}{\text{\# total grasps}}$ over trials where exploration succeeded.

\begin{figure}[t]
\begin{tikzpicture}
\begin{axis}[ymin=0, ymax=5, height=4cm, width=0.45\textwidth, 
  ytick={0,1,...,5}, ytick align=outside, ytick pos=left,
  xtick={0,1,...,6}, xtick align=outside, xtick pos=left,
  xlabel= \footnotesize{\# occlusions},
  ylabel={\footnotesize{Traveled Distance [m]}},
  legend pos=north west,
  legend style={draw=none},
  yticklabel style={font=\small},
  xticklabel style={font=\small}]
\addplot+[
  blue, mark options={blue, scale=0.75}, 
  error bars/.cd, 
    y fixed,
    y dir=both, 
    y explicit
] table [x=x, y=y,y error=error, col sep=comma] {
    x,  y,       error
    0,  0.95,    0.38
    1,  1.80,    0.68
    2,  2.43,    0.77
    3,  3.2,    1.57
    4,  3.42,    1.1
    5,  2.9,    1.05
    
};
\addplot+[
  red, mark options={red, scale=0.75},
  error bars/.cd, 
    y fixed,
    y dir=both, 
    y explicit
] table [x=x, y=y,y error=error, col sep=comma] {
    x,  y,       error
    0,  0.96,    0.44
    1,  1.42,    0.22
    2,  1.40,    0.26
    3,  1.8,     0.37
    4,  1.82,    0.54
    5,  1.57,    0.45
    
};
\addlegendentry{\footnotesize{Baseline}}
\addlegendentry{\footnotesize{\name}}
\end{axis}
\end{tikzpicture}
\vspace{-0.05in}
\caption{\textbf{Traveled Distance.} The figure plots the average traveled distance of the gripper's tip for baseline (blue) and \name\ (red) in successful trials across 0-5 occlusion. Error bars represent the standard deviation.}
\label{fig:travel}
\vspace{-0.15in}
\end{figure}
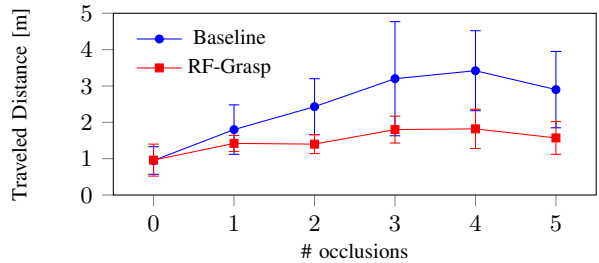

\begin{figure*}[thp]
	\centering
	\begin{tabular}{ccc}
		\epsfig{file=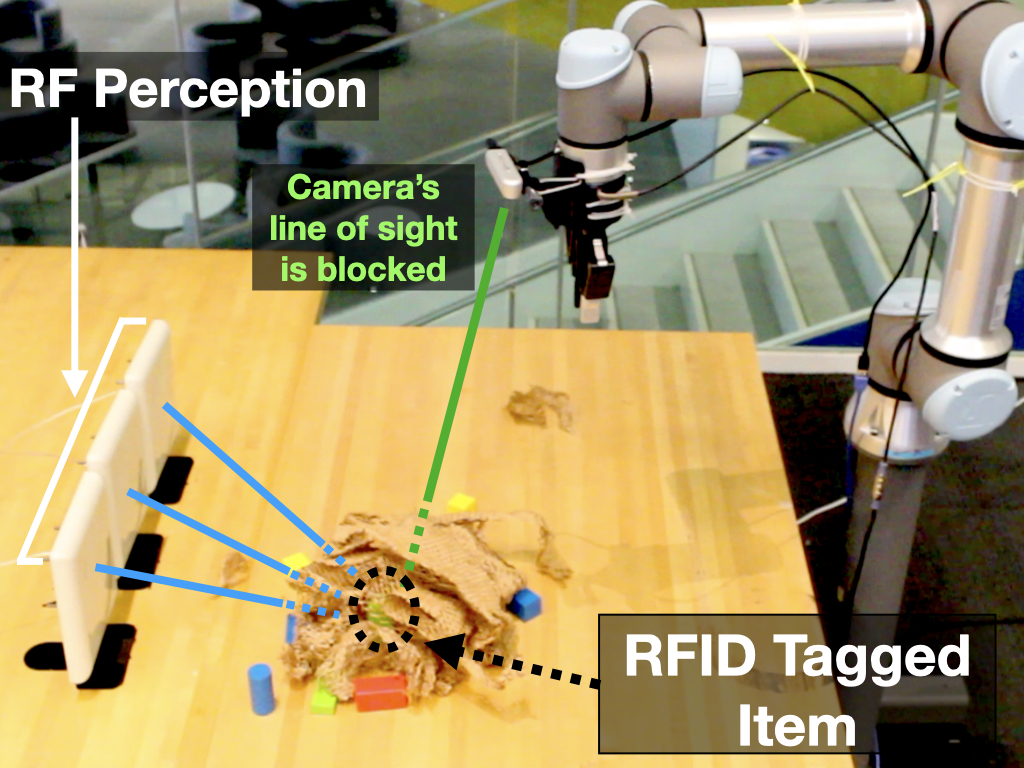,width=0.29\textwidth}
		&
		\epsfig{file=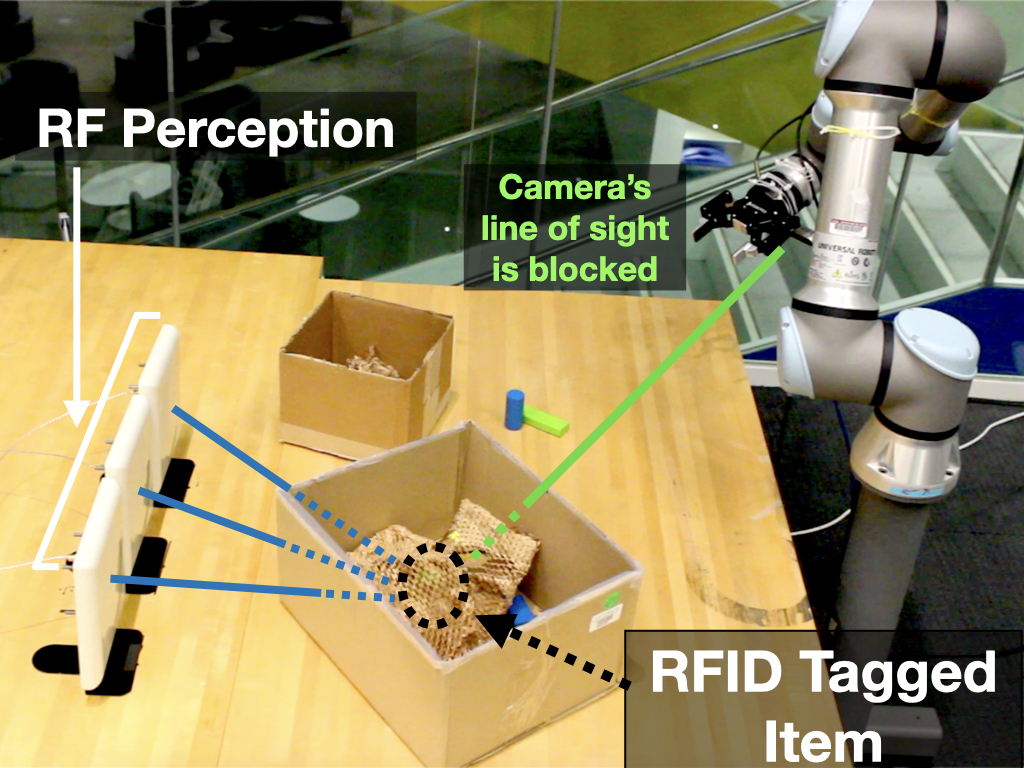,width=0.29\textwidth}
		&
		\epsfig{file=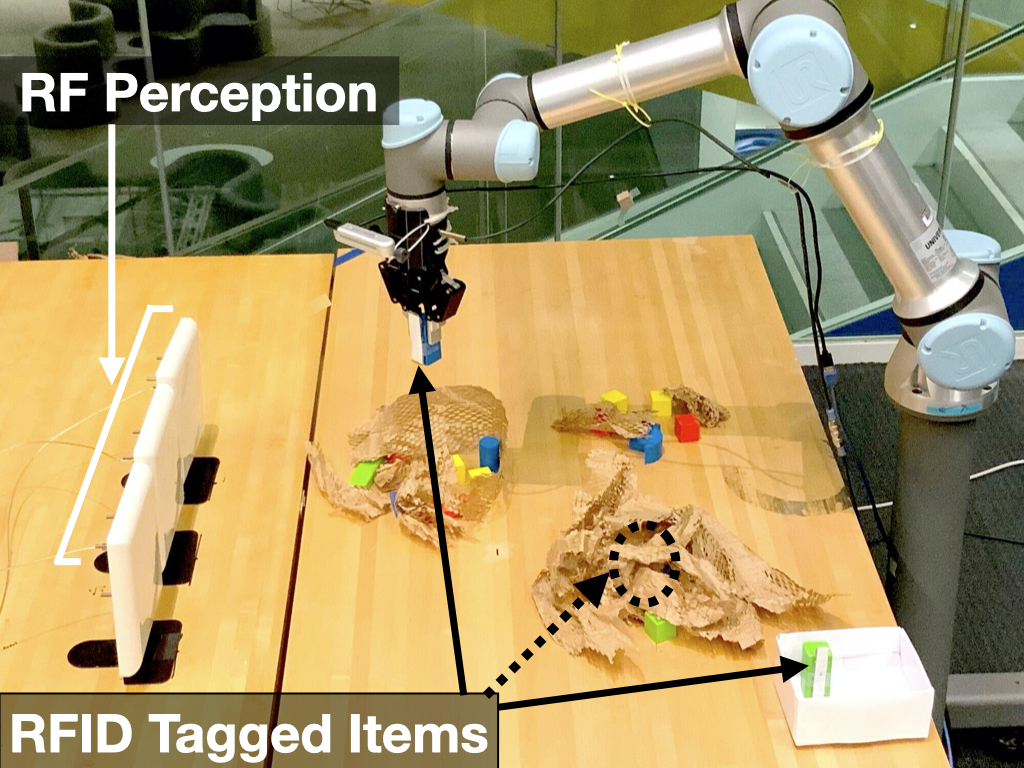,width=0.29\textwidth}
		\\
		
		\footnotesize{(a) Under Soft Cover}
		&
		\footnotesize{(b) Multi-obstacle and Multi-pile }
		&
		\footnotesize{(c) Prioritized Sorting} \vspace{0.2cm}
		\\
\end{tabular}
\vspace{-0.1in}
\caption{\textbf{Generalized Mechanical Search.} We demonatrate \name\ in 3 challenging tasks which the baseline cannot perform.} 
\label{fig:res-qual-fig}
\vspace{-0.15in}
\end{figure*}

\vspace{-0.08in}
\section{Results}
\vspace{-0.1in}
\subsection{Performance Results}
\vspace{-0.05in}
\setulcolor{green}

We evaluated \name\ and the baseline quantitatively by varying the number of large occlusions ($M$) and distractor objects ($N$). Each large occlusion hides a different region of the workspace from the camera. We also varied the initial position of the robot across experimental trials, but ensured the baseline and \name\ shared the same initial position.

\noindent
\textbf{(a) Traveled Distance:} 
We recorded the robot's end-effector position and computed its traveled distance for \name\ and the baseline. We tested six scenarios with 0-5 occluded regions (each with 1-15 distractor objects) and ran 10 trials per system and scenario. 
We placed the robot in an initial pose where the wrist-mounted camera sees one or more occluded regions in its initial observation. Other frontiers and occluded regions were discovered during exploration.

Fig.~\ref{fig:travel} plots the average traveled distance of the gripper for both \name\ and the baseline. For all except $M=0$, our results show that \name\ travels shorter distances (by up to 50\%); moreover, \name\ significantly outperforms the baseline as the number of occlusions increases. This result is expected because \name\ is guided by the RFID's location (in the cost function), which reduces the traveled distance and the time spent on exploring the environment searching for the tagged item in comparison to the baseline. Note that the average traveled distance plateaus beyond four occlusions because the workspace of the robot is limited by its arm size and mobility (and all occluded regions are within that workspace). This result demonstrates the value of \name's first core component (RF-visual servoing) in enabling efficient exploration and trajectory optimization.

\noindent
\textbf{(b) Task Completion Rate:}
Next, we evaluated \name’s completion rate (defined in~\ref{sec:eval}) across the same experiments described above. 
%
Table~\ref{tabCompletion} shows the results. \name\ was able to complete the task in 60 out of 60 trials, while the baseline was unable to complete the task in 7 out of 60 trials. The baseline completion rate decreases as the number of occlusions and complexity in experiments increases.

\begin{table}[h]
\vspace{-0.14in}
\centering
\small
\renewcommand{\arraystretch}{1.25}
\begin{tabular}{lbababa}

& \multicolumn{6}{c}{\textbf{Scenario (M=\#occlusions, N=\#objects)}}\\
\cmidrule{2-7}
\textbf{System}& M=0 N=5 & M=1 N=1 & M=2 N=5 & M=3 N=10 & M=4 N=13 & M=5 N=15\\
\hline
Baseline &    $10_{/10}$  & $10_{/10}$ & $10_{/10}$  & $8_{/10}$  & $7_{/10}$  & $8_{/10}$  \\
RF-Grasp &    $10_{/10}$  &   $10_{/10}$   &  $10_{/10}$ &    $10_{/10}$ &  $10_{/10}$  &  $10_{/10}$ \\
\end{tabular}
\caption{\textbf{Completion Rate.} In each scenario, M denotes number of occlusions, N number of blocks. Results are reported as the number of successfully completed trials out of the total number of experimental trials.}
\label{tabCompletion}
\end{table}

\noindent
\textbf{(c) Grasping Efficiency.} We measured the grasping efficiency across successful trials. \name\ has 78\% efficiency, while the baseline has 68\% efficiency. The improved grasping efficiency for \name\ likely results from variations in lighting which impact color but not RF measurements.

\vspace{-0.08in}
\subsection{Generalized Mechanical Search}
\vspace{-0.06in}
Next, we show that \name\ can successfully perform mechanical search in challenging scenarios where the state-of-the-art baseline is unsuccessful. We consider three such scenarios, shown in Fig.~\ref{fig:res-qual-fig} (see video for demonstration).

\noindent
\textbf{Scenario 1: Under Soft Cover:} We consider scenarios with the target item and more than 20 distractor objects covered with soft package filling sheets (Fig.~\ref{fig:res-qual-fig}(a)). There is no line-of-sight from the wrist-mounted camera or antennas to the item. \name\ localizes the RFID and moves directly above the covered item. 
Because the target is occluded, the robot's attention mechanism biases it to first pick up the cover, but realizes it hasn't picked the target object since the RFID's location has not changed. It puts the cover aside and makes a second grasping attempt. This time, the tracked RFID location changes with the robot's end-effector, and \name\ confirms it has grasped the requested item. \name\ was successful in all 5 trials we performed. The average travelled distance was 2.68m with standard deviation of 0.98. 

In contrast to \name, our baseline is incapable of successfully completing this task because the green-tagged object would remain fully-occluded; thus, it can neither complete its exploration nor efficiently grasp the target object. It is also worth noting that some recent mechanical search system could, in principle, succeed in such a scenario~\cite{ICRA2019search}.

\noindent
\textbf{Scenario 2: Multiple piles and multiple obstacles:}
Next, we tested \name\ with large obstacles and a cover (Fig.~\ref{fig:res-qual-fig}(b)). We used two boxes (with 5 items each) creating two occluded regions.  We placed a tagged item in one box. \name\ was successful in exploring the environment, maneuvering toward the target, removing the cover, and grasping the object. \name\ was successful in all 3 trials we performed. The average travelled distance was 4.13 with standard deviation of 1.91.  To the best of our knowledge, this is a novel task that existing robotic systems cannot perform.

\noindent
\textbf{Scenario 3: Multiple piles and multiple target objects:} Our final scenario involves mechanical search for \textit{all} objects belonging to a semantic class. An example is shown in Fig.~\ref{fig:res-qual-fig}(c), where the robot needs to extract all RFID-tagged items that have a certain feature (e.g., the same dimensions) and sort them into a separate bin. The RF perception module reads and locates all RFID-tagged items, and determines which it needs to grasp. Subsequently, the robot performs mechanical search for each of the items, picks them up, and drops them in the white bin to the bottom right. \name\ succeeded and declared task completion once it localized all target objects to the final bin. The total travelled distance (for 3 objects) was 16.18m. To the best of our knowledge, this is also a novel task that existing systems cannot perform.
\vspace{-0.125in}
\section{Discussion \& Conclusion}
\vspace{-0.1in}
\setulcolor{green}
\name\ fuses RF and visual information to enable robotic grasping of fully-occluded objects. \name\ can be extended in multiple ways to overcome its current limitations: 1) By applying the RF attention to networks capable of grasping more complex items at different angles, this system could be extended to more complex setups. 2) The current system requires separate components for RF localization and grasping. This necessitates a calibration phase for RF-eye-hand coordination. Exploring fully integrated robotic systems could remove this requirement. 3) The underlying approach of this paper can be extended beyond mechanical search to other robotic tasks using RFIDs, including semantic grasping, scene understanding, and human-robot interaction. 

Fundamentally, this work creates a new bridge between robotic manipulation and RF sensing, and we hope it encourages researchers to explore the intersection of these fields.

\vspace{0.03in}
\noindent
\footnotesize{\textbf{Acknowledgments.} 
We thank the anonymous reviewers, the Signal Kinetics group, and Laura Dodds for their feedback. The research is sponsored by the National Science Foundation, NTT DATA, Toppan, Toppan Forms, and Abdul Latif Jameel Water and Food Systems Lab (J-WAFS).}



{
\bibliographystyle{IEEEtran.bst}
\bibliography{paper}

\begin{thebibliography}{10}
\providecommand{\url}[1]{#1}
\csname url@rmstyle\endcsname
\providecommand{\newblock}{\relax}
\providecommand{\bibinfo}[2]{#2}
\providecommand\BIBentrySTDinterwordspacing{\spaceskip=0pt\relax}
\providecommand\BIBentryALTinterwordstretchfactor{4}
\providecommand\BIBentryALTinterwordspacing{\spaceskip=\fontdimen2\font plus
\BIBentryALTinterwordstretchfactor\fontdimen3\font minus
  \fontdimen4\font\relax}
\providecommand\BIBforeignlanguage[2]{{%
\expandafter\ifx\csname l@#1\endcsname\relax
\typeout{** WARNING: IEEEtran.bst: No hyphenation pattern has been}%
\typeout{** loaded for the language `#1'. Using the pattern for}%
\typeout{** the default language instead.}%
\else
\language=\csname l@#1\endcsname
\fi
#2}}

\bibitem{IROS2019Xray}
M.~Danielczuk, A.~Angelova, V.~Vanhoucke, and K.~Goldberg, ``X-ray: Mechanical
  search for an occluded object by minimizing support of learned occupancy
  distributions,'' \emph{IROS}, 2020.

\bibitem{ICRA2019search}
M.~Danielczuk, A.~Kurenkov, A.~Balakrishna, M.~Matl, D.~Wang,
  R.~Mart{\'\i}n-Mart{\'\i}n, A.~Garg, S.~Savarese, and K.~Goldberg,
  ``Mechanical search: Multi-step retrieval of a target object occluded by
  clutter,'' in \emph{2019 International Conference on Robotics and Automation
  (ICRA)}.\hskip 1em plus 0.5em minus 0.4em\relax IEEE, 2019, pp. 1614--1621.

\bibitem{objectfinding}
T.~Novkovic, R.~Pautrat, F.~Furrer, M.~Breyer, R.~Siegwart, and J.~Nieto,
  ``Object finding in cluttered scenes using interactive perception,'' in
  \emph{2020 IEEE International Conference on Robotics and Automation
  (ICRA)}.\hskip 1em plus 0.5em minus 0.4em\relax IEEE, 2020, pp. 8338--8344.

\bibitem{graspinvisible}
Y.~Yang, H.~Liang, and C.~Choi, ``A deep learning approach to grasping the
  invisible,'' \emph{IEEE Robotics and Automation Letters}, vol.~5, no.~2, pp.
  2232--2239, 2020.

\bibitem{ikeabot}
R.~A. Knepper, T.~Layton, J.~Romanishin, and D.~Rus, ``Ikeabot: An autonomous
  multi-robot coordinated furniture assembly system,'' in \emph{2013 IEEE
  International conference on robotics and automation}.\hskip 1em plus 0.5em
  minus 0.4em\relax IEEE, 2013, pp. 855--862.

\bibitem{BerkeleyGraspOcclusion}
G.~Kahn, P.~Sujan, S.~Patil, S.~Bopardikar, J.~Ryde, K.~Goldberg, and
  P.~Abbeel, ``Active exploration using trajectory optimization for robotic
  grasping in the presence of occlusions,'' in \emph{2015 IEEE International
  Conference on Robotics and Automation (ICRA)}.\hskip 1em plus 0.5em minus
  0.4em\relax IEEE, 2015, pp. 4783--4790.

\bibitem{afforfances1}
D.~Katz, A.~Venkatraman, M.~Kazemi, J.~A. Bagnell, and A.~Stentz, ``Perceiving,
  learning, and exploiting object affordances for autonomous pile
  manipulation,'' \emph{Autonomous Robots}, vol.~37, no.~4, pp. 369--382, 2014.

\bibitem{planningclutter}
M.~Dogar, K.~Hsiao, M.~Ciocarlie, and S.~Srinivasa, ``Physics-based grasp
  planning through clutter,'' in \emph{Proceedings of Robotics: Science and
  Systems VIII}, July 2012.

\bibitem{TurboTrack}
Z.~Luo, Q.~Zhang, Y.~Ma, M.~Singh, and F.~Adib, ``3d backscatter localization
  for fine-grained robotics,'' in \emph{16th $\{$USENIX$\}$ Symposium on
  Networked Systems Design and Implementation ($\{$NSDI$\}$ 19)}, 2019, pp.
  765--782.

\bibitem{RFCapture}
F.~Adib, C.-Y. Hsu, H.~Mao, D.~Katabi, and F.~Durand, ``Capturing the human
  figure through a wall,'' \emph{ACM Transactions on Graphics (TOG)}, vol.~34,
  no.~6, p. 219, 2015.

\bibitem{Imping}
``{Impinj R420},'' {\url{http://www.imping.com}}, {Imping Inc.}

\bibitem{RFIDmarket}
R.~Das, ``{RFID Forecasts, Players and Opportunities 2019-2029},'' IDTechx,
  2019.

\bibitem{PinIt}
J.~Wang and D.~Katabi, ``Dude, where's my card? rfid positioning that works
  with multipath and non-line of sight,'' in \emph{{ACM SIGCOMM}}, 2013.

\bibitem{RFCompass}
J.~Wang, F.~Adib, R.~Knepper, D.~Katabi, and D.~Rus, ``{RF-Compass: Robot
  Object Manipulation Using RFIDs},'' in \emph{{ACM MobiCom}}, 2013.

\bibitem{MobiTag}
L.~Shangguan and K.~Jamieson, ``The design and implementation of a mobile rfid
  tag sorting robot,'' in \emph{Proceedings of the 14th annual international
  conference on mobile systems, applications, and services}, 2016, pp. 31--42.

\bibitem{RFind}
Y.~Ma, N.~Selby, and F.~Adib, ``Minding the billions: Ultrawideband
  localization for deployed rfid tags,'' \emph{ACM MobiCom}, 2017.

\bibitem{navigation1}
S.~Park and S.~Hashimoto, ``Autonomous mobile robot navigation using passive
  rfid in indoor environment,'' \emph{IEEE Transactions on industrial
  electronics}, vol.~56, no.~7, pp. 2366--2373, 2009.

\bibitem{navigation2}
W.~Gueaieb and M.~S. Miah, ``An intelligent mobile robot navigation technique
  using rfid technology,'' \emph{IEEE Transactions on Instrumentation and
  Measurement}, vol.~57, no.~9, pp. 1908--1917, 2008.

\bibitem{navigation3}
M.~Kim and N.~Y. Chong, ``Direction sensing rfid reader for mobile robot
  navigation,'' \emph{IEEE Transactions on Automation Science and Engineering},
  vol.~6, no.~1, pp. 44--54, 2008.

\bibitem{T25}
T.~Deyle, C.~J. Tralie, M.~S. Reynolds, and C.~C. Kemp, ``In-hand radio
  frequency identification (rfid) for robotic manipulation,'' in \emph{2013
  IEEE International Conference on Robotics and Automation}.\hskip 1em plus
  0.5em minus 0.4em\relax IEEE, 2013, pp. 1234--1241.

\bibitem{T28}
T.~Deyle, C.~Anderson, C.~C. Kemp, and M.~S. Reynolds, ``A foveated passive uhf
  rfid system for mobile manipulation,'' in \emph{2008 IEEE/RSJ International
  Conference on Intelligent Robots and Systems}.\hskip 1em plus 0.5em minus
  0.4em\relax IEEE, 2008, pp. 3711--3716.

\bibitem{T31}
T.~Deyle, M.~S. Reynolds, and C.~C. Kemp, ``Finding and navigating to household
  objects with uhf rfid tags by optimizing rf signal strength,'' in \emph{2014
  IEEE/RSJ International Conference on Intelligent Robots and Systems}.\hskip
  1em plus 0.5em minus 0.4em\relax IEEE, 2014, pp. 2579--2586.

\bibitem{RFIDfusion}
T.~Deyle, H.~Nguyen, M.~Reynolds, and C.~C. Kemp, ``Rf vision: Rfid receive
  signal strength indicator (rssi) images for sensor fusion and mobile
  manipulation,'' in \emph{2009 IEEE/RSJ International Conference on
  Intelligent Robots and Systems}.\hskip 1em plus 0.5em minus 0.4em\relax IEEE,
  2009, pp. 5553--5560.

\bibitem{PushGrasp}
A.~Zeng, S.~Song, S.~Welker, J.~Lee, A.~Rodriguez, and T.~Funkhouser,
  ``Learning synergies between pushing and grasping with self-supervised deep
  reinforcement learning,'' in \emph{2018 IEEE/RSJ International Conference on
  Intelligent Robots and Systems (IROS)}.\hskip 1em plus 0.5em minus
  0.4em\relax IEEE, 2018, pp. 4238--4245.

\bibitem{MVPclutter}
D.~Morrison, P.~Corke, and J.~Leitner, ``Multi-view picking: Next-best-view
  reaching for improved grasping in clutter,'' in \emph{2019 International
  Conference on Robotics and Automation (ICRA)}.\hskip 1em plus 0.5em minus
  0.4em\relax IEEE, 2019, pp. 8762--8768.

\bibitem{rpp}
A.~Zeng, S.~Song, K.-T. Yu, E.~Donlon, F.~R. Hogan, M.~Bauza, D.~Ma, O.~Taylor,
  M.~Liu, E.~Romo, \emph{et~al.}, ``Robotic pick-and-place of novel objects in
  clutter with multi-affordance grasping and cross-domain image matching,'' in
  \emph{2018 IEEE international conference on robotics and automation
  (ICRA)}.\hskip 1em plus 0.5em minus 0.4em\relax IEEE, 2018, pp. 1--8.

\bibitem{graspocclusion}
S.-K. Kim and M.~Likhachev, ``Planning for grasp selection of partially
  occluded objects,'' in \emph{2016 IEEE International Conference on Robotics
  and Automation (ICRA)}.\hskip 1em plus 0.5em minus 0.4em\relax IEEE, 2016,
  pp. 3971--3978.

\bibitem{semantic1}
E.~Jang, S.~Vijayanarasimhan, P.~Pastor, J.~Ibarz, and S.~Levine, ``End-to-end
  learning of semantic grasping,'' in \emph{Conference on Robot Learning},
  2017, pp. 119--132.

\bibitem{occlusionaware}
X.~Huang, I.~Walker, and S.~Birchfield, ``Occlusion-aware reconstruction and
  manipulation of 3d articulated objects,'' in \emph{2012 IEEE International
  Conference on Robotics and Automation}.\hskip 1em plus 0.5em minus
  0.4em\relax IEEE, 2012, pp. 1365--1371.

\bibitem{price2019inferring}
A.~Price, L.~Jin, and D.~Berenson, ``Inferring occluded geometry improves
  performance when retrieving an object from dense clutter,''
  \emph{International Symposium on Robotics Research (ISRR)}, 2019.

\bibitem{activeperception1}
A.~Aydemir, K.~Sj{\"o}{\"o}, J.~Folkesson, A.~Pronobis, and P.~Jensfelt,
  ``Search in the real world: Active visual object search based on spatial
  relations,'' in \emph{2011 IEEE International Conference on Robotics and
  Automation}.\hskip 1em plus 0.5em minus 0.4em\relax IEEE, 2011, pp.
  2818--2824.

\bibitem{activeperception2}
R.~Bajcsy, ``Active perception,'' \emph{Proceedings of the IEEE}, vol.~76,
  no.~8, pp. 966--1005, 1988.

\bibitem{interactiveperception}
J.~Bohg, K.~Hausman, B.~Sankaran, O.~Brock, D.~Kragic, S.~Schaal, and G.~S.
  Sukhatme, ``Interactive perception: Leveraging action in perception and
  perception in action,'' \emph{IEEE Transactions on Robotics}, vol.~33, no.~6,
  pp. 1273--1291, 2017.

\bibitem{T36}
C.~Nam, J.~Lee, Y.~Cho, J.~Lee, D.~H. Kim, and C.~Kim, ``Planning for target
  retrieval using a robotic manipulator in cluttered and occluded
  environments,'' \emph{arXiv preprint arXiv:1907.03956}, 2019.

\bibitem{T37}
Y.~Cui, J.~Ooga, A.~Ogawa, and T.~Matsubara, ``Probabilistic active filtering
  with gaussian processes for occluded object search in clutter,''
  \emph{Applied Intelligence}, pp. 1--15, 2020.

\bibitem{T38}
K.~Wada, K.~Okada, and M.~Inaba, ``Joint learning of instance and semantic
  segmentation for robotic pick-and-place with heavy occlusions in clutter,''
  in \emph{2019 International Conference on Robotics and Automation
  (ICRA)}.\hskip 1em plus 0.5em minus 0.4em\relax IEEE, 2019, pp. 9558--9564.

\bibitem{T39}
C.~Chen, H.-Y. Li, X.~Zhang, X.~Liu, and U.-X. Tan, ``Towards robotic picking
  of targets with background distractors using deep reinforcement learning,''
  in \emph{2019 WRC Symposium on Advanced Robotics and Automation (WRC
  SARA)}.\hskip 1em plus 0.5em minus 0.4em\relax IEEE, 2019, pp. 166--171.

\bibitem{prioritized}
T.~Schaul, J.~Quan, I.~Antonoglou, and D.~Silver, ``Prioritized experience
  replay,'' \emph{arXiv preprint arXiv:1511.05952}, 2015.

\bibitem{moveit}
{Ettus Research, CDA-2990}, {\url{https://moveit.ros.org/}}.

\bibitem{OMPL}
{The Open Motion Planning Library}, {\url{https://ompl.kavrakilab.org/}}.

\bibitem{yak}
{Yak}, {\url{https://github.com/ros-industrial/yak}}.

\bibitem{PCL}
{The Point Cloud Library (PCL)}, {\url{https://pointclouds.org/}}.

\bibitem{forces}
{Forces Pro}, {\url{https://www.embotech.com/}}.

\bibitem{Alien}
{A}lien~{T}echnology {I}nc., ``{ALN-9640 Squiggle Inlay},''
  www.alientechnology.com.

\bibitem{vrep}
{VREP}, {\url{https://www.coppeliarobotics.com/}}.

\bibitem{gazebo}
{Gazebo}, \url{http://gazebosim.org/}.

\bibitem{rviz}
{Rviz}, {\url{http://wiki.ros.org/rviz}}.

\end{thebibliography}
}
\end{sloppypar}
\end{document}